\documentclass[letterpaper, 10 pt, conference]{ieeeconf}  

\IEEEoverridecommandlockouts                              
\usepackage{graphicx}
\usepackage{amsfonts}
\usepackage{multirow} 

\usepackage{color}
\usepackage{amsmath}
\usepackage{mathrsfs}
\usepackage{makecell}
\usepackage{url}
\usepackage{array}
\usepackage{booktabs}
\usepackage{balance}
\usepackage{bbding}
\usepackage{tabularx}
\usepackage{pifont}

\definecolor{xf}{RGB}{69,137,148}
\definecolor{sf}{RGB}{0,0,205}

\newcommand{\xf}[1]{{\color{black} #1}}
\newcommand{\ssf}[1]{{\color{black} #1}}

\newcommand{\edit}[1]{{\color{black} #1}}
\newcommand{\reff}[1]{{\color{black} #1}}
\title{\LARGE \bf
Preparation of Papers for IEEE Sponsored Conferences \& Symposia*
}
\author{Fei Sheng$^{1,\dag}$, Feng Xue$^{1,\dag}$, Yicong Chang$^{1}$, Wenteng Liang$^{1}$ and Anlong Ming$^{1,*}$
\thanks{\dag Equal contribution.}
\thanks{*Corresponding author.}
\thanks{$^{1}$Beijing University of Posts and Telecommunications, Beijing, China,
		{\tt\small \{shengfei,xuefeng,yicongchang,liangwenteng,mal\}
		@bupt.edu.cn}}%
\edit{\thanks{This work was supported by the national key R \& D program intergovernmental international science and technology innovation cooperation project 2021YFE0101600, and Excellent Ph.D. Students Foundation CX2020114.}}}

\begin{document}

\title{\LARGE \bf Monocular Depth Distribution Alignment with Low Computation}

\maketitle
\thispagestyle{empty}
\pagestyle{empty}
\begin{abstract}
The performance of monocular depth estimation generally depends on the amount of parameters and computational cost.
It leads to a large accuracy contrast between light-weight networks and heavy-weight networks,
which limits their application in the real world.
In this paper,
we model the majority of accuracy contrast between them as the difference of depth distribution,
which we call 'Distribution drift'.
To this end,
a distribution alignment network (DANet) is proposed.
We firstly design a pyramid scene transformer (PST) module to capture inter-region interaction in multiple scales.
By perceiving the difference of depth features between every two regions,
DANet tends to predict a reasonable scene structure,
which fits the shape of distribution to ground truth.
Then, we propose a local-global optimization (LGO) scheme to realize the supervision of global range of scene depth.
Thanks to the alignment of depth distribution shape and scene depth range,
DANet sharply alleviates the distribution drift,
and achieves a comparable performance with prior heavy-weight methods,
but uses only 1\% floating-point operations per second (FLOPs) of them.
The experiments on two datasets,
namely the widely used  NYUDv2 dataset and the more challenging iBims-1 dataset,
demonstrate the effectiveness of our method.
The source code is available at \url{https://github.com/YiLiM1/DANet}.
\end{abstract}

\section{Introduction}
Monocular depth estimation (MDE) aims to infer the 3D space for a given 2D image,
which has been widely applied in many computer vision and robotics tasks,
e.g.,
visual SLAM \cite{MVO,CNNSLAM,PseudoRGBD},
monocular 3D object detection \cite{M3ODPL,2020Monocular}, 
obstacle avoidance \cite{JMOD2,MVOA,Tiny_Obstacle1},
and augmented reality \cite{CVDE}.
\xf{They raise high demand for both the accuracy and speed of MED.}

With the development of deep learning,
many impressive works \cite{ Laina, DORN_2018_CVPR, LookDeeper, Midas} have emerged,
most of which focus on improving accuracy of predicted depth.
However,
when we attempt to reduce the parameter and computation of these models 
their accuracy drops sharply.
The degradation is mainly caused by the inadequate feature representation of pixel-wise continuous depth value. 
This phenomenon also appears in some recent algorithms \cite{Jointdepth,Fastdepth} that realize real-time MDE.
To our knowledge, it is difficult for the prior methods to run in low latency while achieving a similar performance as the networks focusing on accuracy.

\begin{figure}
\centering
\includegraphics[width=1 \linewidth]{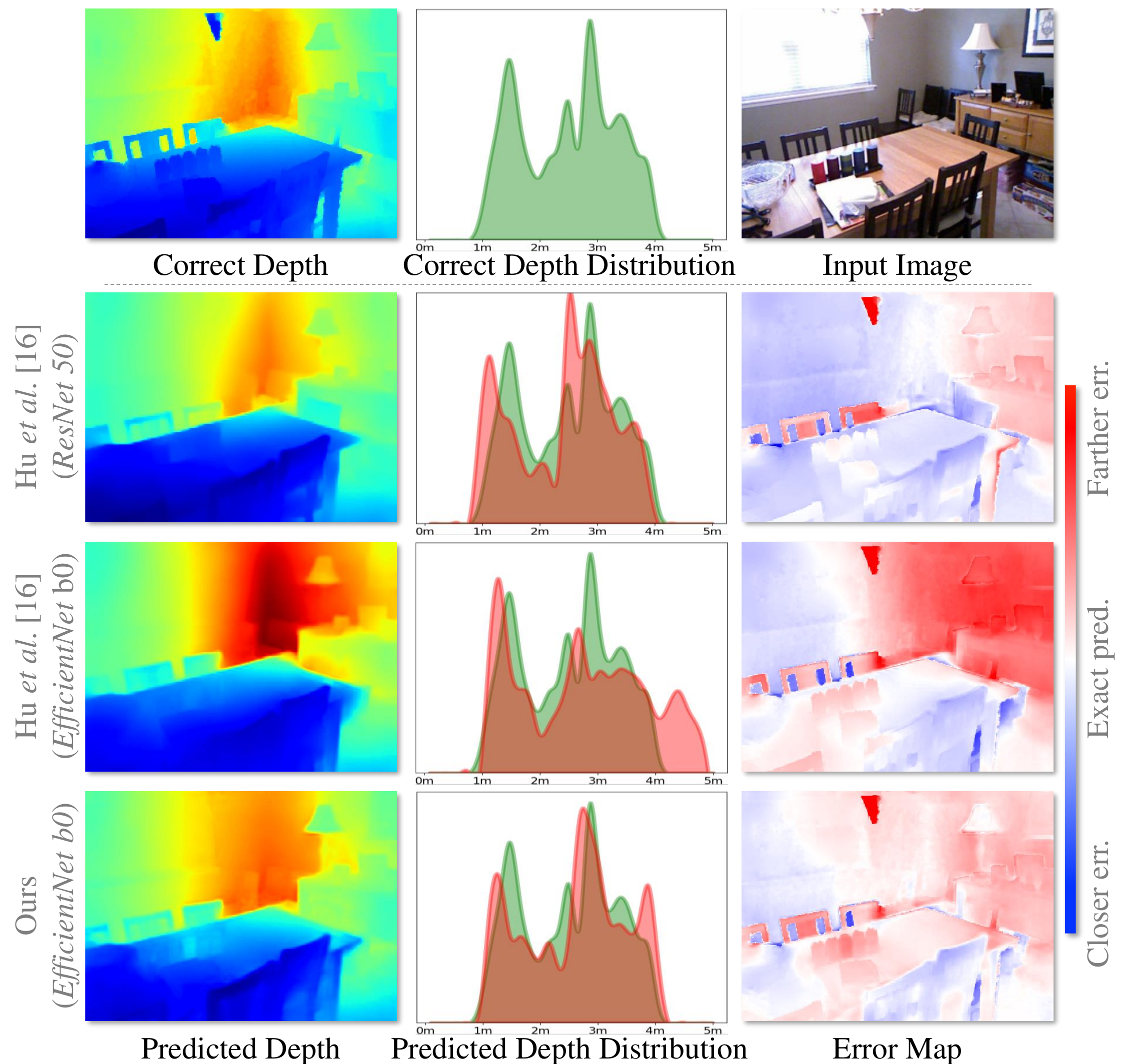}
\caption{Illustration of distribution drift phenomenon.
The depth distribution is represented by the histogram of depth values,
green for correct depth and red for predicted depth.
The error map describes the pixel-wise error of depth, 
with red indicating too far and blue indicating too close.}
\label{fig:intro}
\vspace{-0.3cm} 
\end{figure}

The motivation of this paper is the \ssf{observed} major degradation of the light-weight MDE models compared to heavy-weight MDE models.
We found that for the light-weight MDE networks,
there are usually whole pieces of pixel in the prediction that are smaller or larger than the correct depth monolithically,
which is the main indicator of accuracy degradation.
As shown in the second error map of \cite{Revisiting} in Fig. \ref{fig:intro},
almost all pixels on the wall are predicted farther,
which can be observed more intuitively in the depth distribution.
Depth distribution shows the proportion of pixels with different depth values.
The light-weight MDE models tend to get a completely different depth distribution from the ground truth,
which is reflected in two differences,
i.e., the shape of depth 
\ssf{distribution} and the full depth range.
We call this issue ‘Distribution Drift’.
As shown in Fig. \ref{fig:intro},
\cite{Revisiting} using a light-weight backbone obtains a different shape \ssf{of} depth distribution and depth range from ground truth.


\begin{figure*}
\vspace{0.1cm} 
 \centering
 \includegraphics[width=1\linewidth]{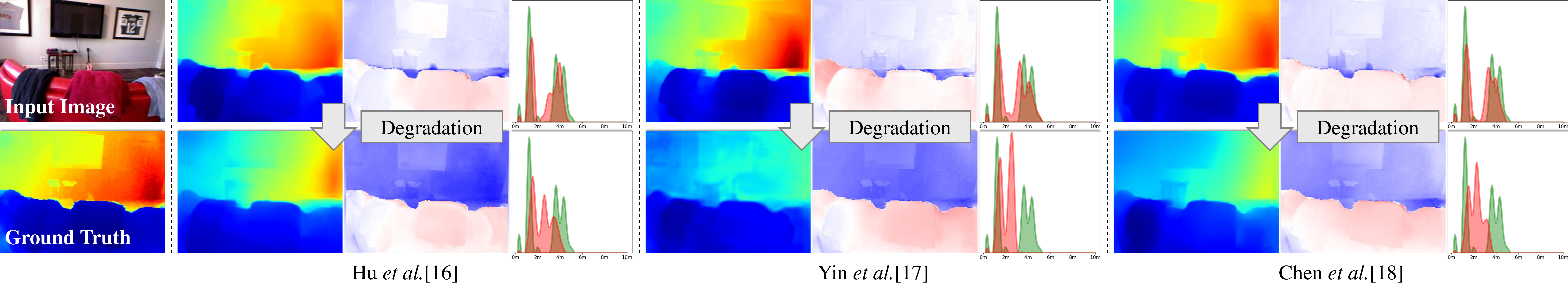}
  \caption{Cases of accuracy degradation 
  \ssf{of} the prior state-of-the-art methods (\cite{Revisiting,Yin_2019_ICCV,SARPN} from left to right).
  For each method,
  the first row shows the prediction, error map and depth distribution of the models using the heavy-weight backbone,
  and the second row for the light-weight backbone.}
 \label{fig:relatedwork}
  \vspace{-0.3cm} 
\end{figure*}

In this paper,
we propose a distribution alignment network to alleviate the distribution drift,
making our method to achieve the performance comparable to the state-of-the-art methods,
while with low latency.
Firstly,
to address the shape deviation of depth distribution,
we propose a pyramid scene transformer (PST).
Since the light-weight models are limited in network depth,
they only extract depth cues in short range.
However, minimal depth changes in a short range can hardly be perceived,
which causes the wrong predicted depth of the whole slice.
In the proposed PST,
we capture the long-range interaction between every two regions in multiple scales,
which constrains the depth relationship between different regions.
Thus PST is beneficial to realize a reliable scene structure.
Then,
to align the depth range,
a local-global optimization (LGO) scheme is proposed to optimize the local depth value and the global depth range simultaneously.
\ssf{By using maximum and minimum depth as supervision,}
the value range of the scene depth is estimated to be aligned with the ground truth.
Experiments prove that we indeed align the distribution of the scene depth,
which helps our method to achieve comparable performance with state-of-the-art methods on NYUDv2 and iBims-1 datasets.

The main contributions of this work lie in:
\begin{itemize}
\item
The distribution drift is studied to reveal the major degradation of light-weight models,
which inspires us to propose a distribution alignment network (DANet).
The DANet exceeds all prior light-weight works,
and achieves a comparable accuracy with heavy-weight models but uses only 1\% FLOPs of them.

\item A pyramid scene transformer (PST) \ssf{module} is proposed to gain long-range interaction between multi-scale regions,
helping DANet to alleviate the shape deviation of predicted depth \ssf{distribution}.

\item A local-global optimization (LGO) scheme is proposed to jointly supervise the network with local depth value and global depth statistics.
\end{itemize}

\section{Related work}
\subsection{Monocular Depth Estimation}
\xf{Several early monocular depth estimators utilize the handcrafted features to estimate depth \cite{Mark3D,Discrete}
but suffer from insufficient expression ability.}
Recently, many CNN-based methods achieve great performance gain.
Eigen \emph{et al.} \cite{Eigen_2015_ICCV,Eigen} propose a coarse-to-fine CNN to estimate depth. 
Laina \emph{et al.} \cite{Laina} propose the up-projection for MDE \xf{to achieve higher accuracy}.
Xue \emph{et al.} \cite{BSNet} improve the boundary accuracy of the predicted depth by a boundary fusion module.
Yin \emph{et al.} \cite{Yin_2019_ICCV} enforce geometric constraints of virtual normal for depth prediction.
Different from these works,
Fu \emph{et al.} \cite{DORN_2018_CVPR} define MDE as a classification task. 
They divide the depth range into a set of bins with a predetermined width. 
Bhat \emph{et al.} \cite{AdaBins} compute bins adaptively for each image.

However,
these prior methods focus heavily on achieving high accuracy with the cost of complexity and runtime,
because they construct a large number of convolutions to obtain sufficient feature representation.
\xf{Fig. \ref{fig:relatedwork} shows three models to verify the issue we find.}
Once the network capacity drops,
they suffer a sharp degradation in accuracy which limits their applications. 
\ssf{In this paper,
we solve this problem by aligning the depth distribution,
thus our method achieves a better trade-off between accuracy and computation.}

\begin{figure*}
\vspace{0.1cm} 
 \centering
 \includegraphics[width=0.95\linewidth]{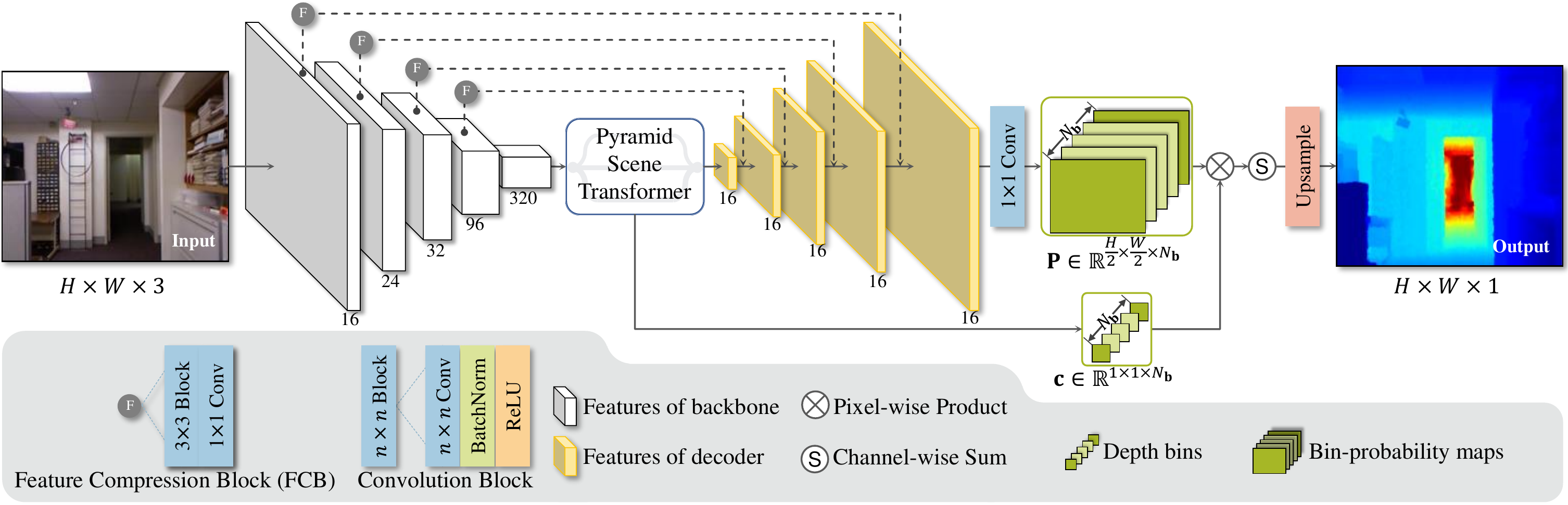}
  \caption{Network architecture of DANet which consists of an encoder-decoder network, pyramid scene transformer.
  The pyramid scene transformer is between the encoder and the decoder,
  which predicts the center of depth bin to combine with the output of decoder.
  }
\label{fig:pipeline}
\vspace{-0.3cm} 
\end{figure*}

\subsection{Real-time Monocular Depth Estimation Methods}
To reduce latency in inference,
several methods are proposed for MDE in recent years.
Wofk \emph{et al.} \cite{Fastdepth} design an extremely light-weight network.
It adopts the MobileNet \cite{MobileNet} and depth-wise separable convolution \cite{Xception} to build the whole network,
followed by network pruning to further reduce computation.
Nekrasov \emph{et al.} \cite{Jointdepth} boost depth estimation by learning semantic segmentation and distilling structured knowledge from large model to light-weight model.
However,
due to the limited capacity of the model,
distribution drift still appears in these light-weight networks.


\subsection{Context Learning}
\reff{Context plays an important role in computer vision tasks \cite{PSPNet,TextTracking,NIPS2012_3e313b9b,YuZhou-IJCV2016-SFVT,ZhouYu}.}
Zhao \emph{et al.} \cite{PSPNet} propose the pyramid pooling to aggregate global context information.
Lu \emph{et al.} \cite{OcclusionEdge2} propose the multi-rate context learner to capture image context by dilated convolution.
Vaswani \emph{et al.} \cite{Transformer} design the transformer to obtain global context by self-attention,
which is used in multiple vision tasks \cite{Vit,AIT,TT}.
This paper proposes a pyramid scene transformer to capture context interaction between multi-scale regions. 





\section{Problem Formulation}
In contrast to other 
methods \cite{Revisiting,SARPN}, following \cite{AdaBins}, the depth range ($[0,10]$ in NYUDv2 dataset) of the whole scene is divided into $N_\mathbf{b}$ bins, and the goal of depth estimation is formulated as \ssf{follow}: for an input image $I\in \mathbb{R}^{H\times W\times 3}$, two tensors are jointly predicted, namely, the center values of depth bins $\mathbf{c} \in \mathbb{R}^{1\times1\times N_\mathbf{b}}$, and the bin-probability maps $\mathbf{P} \in \mathbb{R}^{H\times W \times N_\mathbf{b}}$ indicating the probability of each pixel falling into the corresponding depth bin. In the final predicted depth map, denoted  as $\mathcal{Y}$, each pixel can be formulated as the linear combination of bin probabilities and the bin centers. 
\begin{equation}
    y_i = \sum\nolimits_{n=1}^{N_\mathbf{b}} \mathbf{P}_i(n) \mathbf{c}(n)
    \label{eq:linearcombine}
\end{equation}
where $y_i$ denotes the $i$-th pixel in the prediction $\mathcal{Y}$.
\section{Methodology}
The first subsection outlines the whole architecture of DANet.
The second subsection illustrates the pyramid scene transformer (PST),
and the following subsection presents the local-global optimization (LGO) scheme for aligning the depth range of the scene to the correct range.

\subsection{Network Structure}
Fig. \ref{fig:pipeline} illustrates the architecture of DANet that consists of an encoder,
a pyramid scene transformer,
and a decoder.
\xf{Given an image $I$,
a light-weight backbone EfficientNet B0 \cite{Efficientnet} is used to extract features.}
Assuming that the $i$-th level feature map of the backbone is denoted as $\mathbf{x}_i \in \mathbb{R}^{\frac{H}{2^i}\times\frac{W}{2^i}\times C_i}, (i \in \{1,2,3,4,5\})$,
where $C_i$ is the channel number.
The feature compression block (FCB),
\edit{composed of a $3\times3$ convolution and a $1\times1$ convolution},
is used to reduce the channel number of feature $\mathbf{x}_i$ ($i\in\{1,2,3,4\}$) to 16.
The four FCBs provide multi-level scene \ssf{detail information} with low time cost.
At the end of the encoder,
the PST
is employed to capture the interaction between multi-scale regions from $\mathbf{x}_5$,
meanwhile predict the center values of \ssf{the} depth bin,
i.e., $\mathbf{c}$,
in the scene (see Section \ref{sec:pst}).
\xf{In the decoder,}
four up-scaling stages are employed to gradually enlarge the resolution from $\frac{H}{32}\times\frac{W}{32}$ to $\frac{H}{2}\times\frac{W}{2}$.
Each stage upsamples the last stage output,
and sums it with the same-size feature given by FCB.
Then,
a residual structure containing three $3\times 3$ convolutions is used to fuse these features.
The channel numbers of features in the decoder are set to 16 to meet requirements of low latency and light weight.
At the end of the decoder,
we use a $1\times1$ convolution to learn $N_\mathbf{b}$-dimensional bin-probability map $\mathbf{P}$ from the ﬁnest resolution feature.
Referring to Eq. \ref{eq:linearcombine},
the final prediction $\mathcal{Y}$ is obtained by linear combination of $\mathbf{P}$ and $\mathbf{c}$.

\begin{figure}[t]
\centering
\includegraphics[width=1\linewidth]{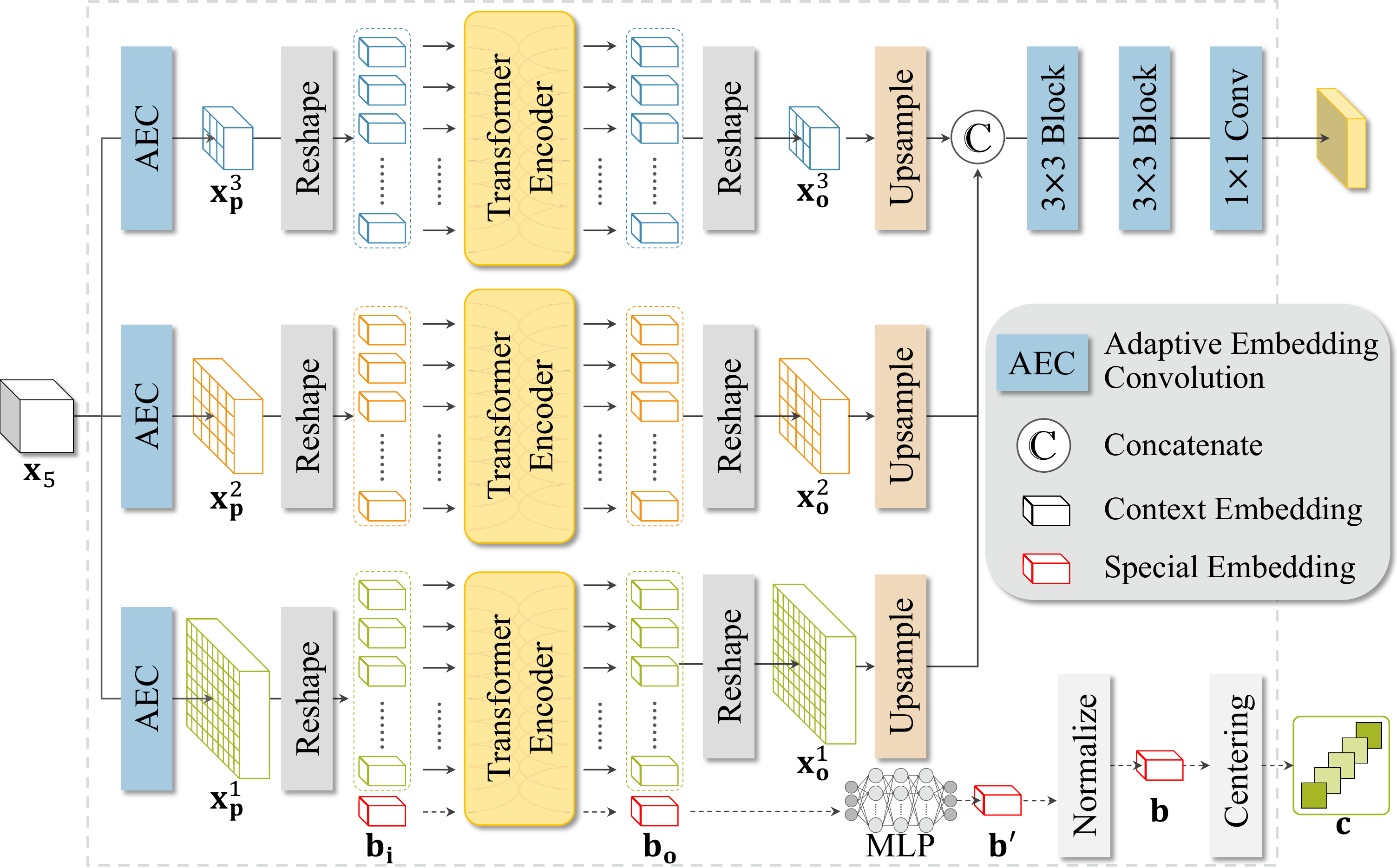}
\caption{The detailed structure of pyramid scene transformer.}
\label{fig:pst}
\vspace{-0.3cm}
\end{figure}

\subsection{Pyramid Scene Transformer}
\label{sec:pst}

\xf{Context interaction models the inter-region relationship of depth features,
which helps to correctly estimate depth difference between regions.
In the global view of the scene,
it plays a significant role in suppressing the shape deviation of the depth distribution.
}To this end,
we design 
the PST to capture context interaction,
which consists of three \edit{independent} parallel paths, as shown in Fig. \ref{fig:pst}.
These paths divide the scene into various-size patches,
respectively,
to cover various-size scene components.
And the relationship between every two patches is captured by a transformer structure \cite{Vit}.

Specifically,
an adaptive embedding convolution (AEC) is firstly designed to gain the multi-scale context embeddings adaptively.
Given the input feature resolution $H_\mathbf{i}\times W_\mathbf{i}$ and the expected output resolution $H_\mathbf{o}\times W_\mathbf{o}$,
AEC is defined as:
\begin{itemize}
\item Stride in X direction: $s_\mathbf{x} = \lfloor W_\mathbf{i} / W_\mathbf{o}\rfloor$
\item Stride in Y direction: $s_\mathbf{y} = \lfloor H_\mathbf{i} / H_\mathbf{o}\rfloor$
\item Kernel size: $\big(H_\mathbf{i}-s_\mathbf{y}(H_\mathbf{o}-1)\big)\times\big(W_\mathbf{i}-s_\mathbf{x}(W_\mathbf{o}-1)\big)$
\end{itemize}
By using AEC,
these paths re-scale $\mathbf{x}_5$ into tensors with three sizes:
\xf{$\mathbf{x}^j_{\mathbf{p}} \in \mathbb{R}^{\frac{H}{2^{j+4}}\times\frac{W}{2^{j+4}}\times C_e}$,
where $j \in \{1,2,3\}$ is the path number.}
Each pixel in $\mathbf{x}^j_{\mathbf{p}}$ represents the $C_e$-dimensional context embedding of a patch in the scene.
Secondly, in a path, all embeddings are fed into a transformer encoder after adding a 1-D learned positional encoding \cite{Vit}. 
\edit{The transformer encoder is utilized to perceive the interaction between every two embeddings, and output a sequence of embeddings with the same size as the input embeddings.}
Note that the first path is different from the two others. It appends an additional $C_e$-dimensional embedding $\mathbf{b}_\mathbf{i}$ together with context embeddings, and outputs a special embedding $\mathbf{b}_\mathbf{o}$ which has the same size as $\mathbf{b}_\mathbf{i}$.
Thirdly,
in each path,
the output embeddings are reshaped to build a tensor $\mathbf{x}^i_{\mathbf{o}}, i\in\{1,2,3\}$ which has the same size as $\mathbf{x}^i_{\mathbf{p}}, i\in\{1,2,3\}$.
Then,
all output tensors $\mathbf{x}^i_{\mathbf{o}}$ are upsampled to $\frac{H}{32}\times\frac{W}{32}\times C_e$,
so that they can be concatenated.
\edit{The concatenated feature is then compressed to 16 channels through two $3\times3$ convolution and a $1\times1$ convolution,
and fed into the decoder.}
Meanwhile,
in the first path,
\edit{the output special embedding $\mathbf{b}_\mathbf{o}$ is fed into a multi-layer perceptron to obtain a $N_\mathbf{b}$-dimensional vector.}
Subsequently,
in the same way as \cite{AdaBins},
the vector $\mathbf{b}'$ is normalized to obtain the depth-range widths vector $\mathbf{b}$:
$b_i = \frac{b'_i+\tau}{\sum^{N_{\mathbf{b}}}_{j=1}(b'_j +\tau)}, i\in\{1,2,...,N_\mathbf{b}\}$.
And the center of \ssf{bin}  $\mathbf{c}$ is obtained as follows:
$c_i = d_{min}+(d_{max}-d_{min})(b_i/2+\sum_{j=1}^{i-1}b_j)$,
where $d_{min}, d_{max}$ are the minimum and maximum depth values,
and $c_i$ is the $i$-th value in $\mathbf{c}$.

Since the transformer extracts the context interaction of each two patches in a scene,
each output embedding encodes the depth interaction from one patch to all other patches.
And different paths correspond to the depth correlation of patch in various scales.
\edit{Moreover,
unlike \cite{AdaBins}, PST is between the encoder and decoder to minimize the amount of computation.}



\subsection{Local-Global Optimization for Depth Range Learning}
\xf{To align the global depth range,}
we propose a local-global optimization (LGO) scheme, 
\xf{which trains DANet by two stages.}
In the local stage,
we perform two local errors referred from \cite{AdaBins} as supervision.
In the global stage,
\xf{we propose min-max loss and range-based pixel weight to learn the global depth range and optimize the whole depth.}


\subsubsection{Loss of local stage}
\xf{The local stage aims to optimize the pixel-wise depth.
To this end,
a scaled version of the Scale-Invariant (SSI) loss \cite{bts} is used to minimize the pixel-wise error between the predicted depth and correct depth:}
\begin{equation}
    L_{pixel}=\sqrt{\frac{1}{N}\sum\nolimits_{i}^{}h_i^2-\frac{u}{N^2}\left(\sum\nolimits_{i}h_i \right)^2}
    \label{eq:L_pixel}
\end{equation}
where $h_i=\left(\lambda_i+1\right)\cdot\left(\log{}{y_i}-\log{}{g_i}\right)$,
and $N$ is the pixel number of an image.
$y_{i}, g_{i}$ are the predicted and correct depth respectively.
$\lambda_i$ is a weight parameter of pixel $i$.
In the local stage,
$\lambda_i$ is set to $0$.
\edit{Furthermore, 
following \cite{AdaBins},
the bi-directional Chamfer Loss \cite{AdaBins} is employed as a regularizer to optimize the bin centers $\mathbf{c}$ to be close to the ground truth.}
\begin{equation}
    L_{bin}=\sum_{x\in \mathcal{X}} \min_{c_i \in \mathbf{c}} \left \|x-c_i\right \|^2 +\sum_{c_i\in \mathbf{c}} \min_{x\in \mathcal{X}} \left \|x-c_i\right \|^2
    \label{eq:L_bin}
\end{equation}
where $\mathcal{X}$ is the set of all depth values in the ground truth.


\xf{
\subsubsection{Loss of global stage}
The global stage aims to learn the depth range.
In this stage,
we supervise the first and last bin center in $\mathbf{c}$ by a new designed min-max loss:}
\begin{equation}
    L_{minmax}=\left\|c_{1}-min(g)\right\|_1+\left\|c_{N_\mathbf{b}}-max(g)\right\|_1
    \label{eq:L_minmax}
\end{equation}
where $c_i$ is $i$-th value of $\mathbf{c}$.
$min$ and $max$ are \ssf{the} operation of taking minimum and maximum value, respectively.
The min-max loss affects all pixels during back-propagation by supervising the bins, \ssf{so that it squeezes all predicted depth values into range $[g_{min},g_{max}]$.} 


\begin{figure}[t]
\vspace{0.1cm} 
\centering
\includegraphics[width=1\linewidth]{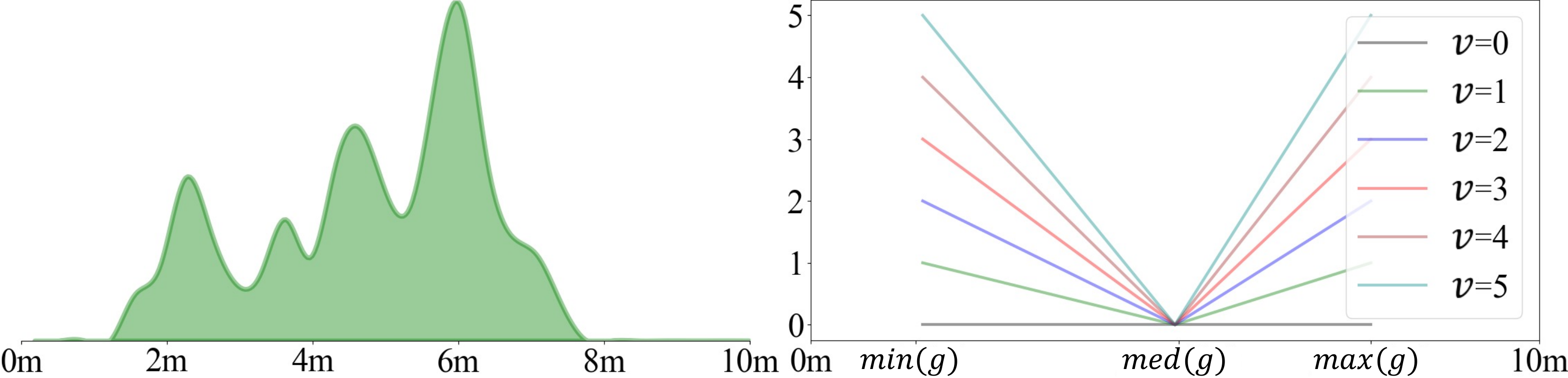}
\caption{The distribution of a depth map and value of $\lambda_i^D$ with different $v$.} 
\label{fig:valley_a}
\vspace{-0.4cm} 
\end{figure}

However,
since the amount of pixels with the largest and smallest depth is small in scenes,
the network might be insensitive to these pixels.
Thus, we additionally assign a depth-related weight to each pixel. 
In the global stage,
the parameter $\lambda^{D}_i$ in Eq. \ref{eq:L_pixel} is taken as the depth-related weight of a pixel,
which is proportional to the difference from the pixel's depth to the median depth value in ground truth.
\begin{equation}
    \lambda^{D}_i=\left\{
    \begin{array}{rcl}
    \frac{v(med(g)-g_i)}{med(g) - min(g)}& & if\,g_i\leq med(g)\\
    \frac{v(g_i-med(g))}{max(g) - med(g)}& & otherwise\\
    \end{array} \right.
    \label{eq:labdaD}
\end{equation}

where $i$ is pixel index,
and $v$ is a coefficient.
$med$ denotes the operation of taking medium value.
As shown in Fig. \ref{fig:valley_a},
if the correct depth $g_i$ is close to $max(g)$ or $min(g)$,
the $\lambda^{D}_i$ is close to $v$,
which means that the network pays more attention \ssf{to} this pixel $i$.
If the correct depth $g_i$ is close to $med(g)$,
the $\lambda^{D}_i$ tends to be $0$.
In this way,
DANet pays more attention to pixels with small and large depth,
and predicts a more reasonable depth range.




\subsubsection{Training scheme}
Combined with the min-max loss and $\lambda^{D}_i$,
the total loss is formulated as:
\begin{equation}
    L_{total}=\alpha L_{pixel}+\beta L_{bin}+ \gamma L_{minmax} 
    \label{eq:L_total}
\end{equation}
where $\alpha,\beta,\gamma$ are hyper-parameters.
In the first stage,
we set $\alpha=10$, $\beta=0.1$, $\gamma=0$, $u=0.85$ and $\lambda_i=0$.
The first stage optimizes the pixel-wise depth preliminarily.
In the second stage,
we set $\alpha=10$, $\beta=0.1$, $\gamma=0.1$, $u=0.85$, $v=1$ and $\lambda_i=\lambda^{D}_i$.
The second stage further optimizes the depth range based on the learned weight of the first stage.




\begin{table*}[t]
\centering
\vspace{0.14cm} 
\caption{Comparisons on NYUDv2 dataset.
Group \ding{172} contains non-lightweight methods.
Group \ding{173} contains light-weight methods.
Group \ding{174} contains the re-implemented models using a same backbone with our method.}
\label{table1}
\begin{tabular}{|c|l|c|c|c|c|c c c|c c c|}
\hline
Groups & Methods & Backbone & Resolution & FLOPs& Params& REL $\downarrow$ & RMS $\downarrow$ & log10 $\downarrow$ & $\delta_1$ $\uparrow$ & $\delta_2$ $\uparrow$ & $\delta_3$ $\uparrow$ \\
\hline
\hline
\multirow{10}{*}{\ding{172}}&Eigen \emph{et al.}\cite{Eigen}&VGG16& $240\times 320$ &31G &240M& 0.215 & 0.772 & 0.095& 0.611 & 0.887 & 0.971 \\
&Eigen \emph{et al.} \cite{Eigen_2015_ICCV}&VGG16&$228\times 304$&23G&-&0.158&0.565& -&0.769&0.950&0.988\\
&Laina \emph{et al.} \cite{Laina}&ResNet50&$240\times 320$&17G&63M&0.127&0.573&0.055&0.811&0.953&0.988\\
&Fu \emph{et al.} \cite{DORN_2018_CVPR}&ResNet101&$240\times 320$&102G&85M&0.118&0.498&0.052&0.828&0.965&0.992\\
&Lee \emph{et al.} \cite{Lee_2019_CVPR}&DenseNet161&$224\times 224$&96G&268M&0.126&0.470&0.054&0.837&0.971&0.994\\
&Hu \emph{et al.} \cite{Revisiting}&ResNet50&$228\times 304$&107G&67M&0.130&0.505&0.057&0.831&0.965&0.991\\
&Chen \emph{et al.} \cite{SARPN}&SENet154&$228\times 304$&150G&258M&0.111&0.420&0.048&0.878&0.976&0.993\\
&Yin \emph{et al.} \cite{Yin_2019_ICCV}&ResNet101&$384\times 384$&184G&90M&0.105&0.406&0.046&0.881&0.976&0.993\\
&Lee \emph{et al.} \cite{bts}&ResNet101&$416 \times 544$&132G&66M&0.113&0.407&0.049&0.871&0.977&0.995\\
&Bhat \emph{et al.} \cite{AdaBins}&EfficientNet b5&$426\times 560$&186G&77M&0.103&0.364&0.044&0.902&0.983&0.997\\
\hline
\multirow{4}{*}{\ding{173}}&Wofk \emph{et al.} \cite{Fastdepth}&MobileNet& $224 \times 224$ & 0.75G&3.9M& 0.162 & 0.591 & -& 0.778& 0.942 & 0.987  \\
&Nekrasov \emph{et al.} \cite{Jointdepth}&MobileNet v2& $480 \times 640$ & 6.49G &2.99M& 0.149 & 0.565 & - & 0.790 & 0.955&0.990 \\
&Yin \emph{et al.} \cite{Yin_2019_ICCV}&MobileNet v2& $338 \times 338$ & 15.6G&2.7M& 0.135 & - &0.060 & 0.813 & 0.958 & 0.991 \\
&Hu \emph{et al.} \cite{boosting}&MobileNet v2& $228 \times 304$ & -&1.7M& 0.138 & 0.499 & 0.059& 0.818 & 0.960 & 0.990   \\
\hline
\multirow{3}{*}{\ding{174}}&Hu \emph{et al.} \cite{Revisiting} \dag &EfficientNet b0& $228 \times 304$ & 14G & 5.3M& 0.142 & 0.505 & 0.059& 0.814 & 0.961 & 0.989 \\
&Chen \emph{et al.} \cite{SARPN} \dag &EfficientNet b0& $228 \times 304$ &8.22G&12M& 0.135 & 0.514 & -  &0.828& 0.963 & 0.990 \\
&Yin \emph{et al.} \cite{Yin_2019_ICCV} \dag&EfficientNet b0& $384 \times 384$ &18G&4.6M& 0.145 & 0.567 & 0.067&0.771& 0.947 & 0.988  \\
\hline
\hline
& Ours &EfficientNet b0& $228\times 304$ &1.5G&8.2M&0.135&0.488&0.057&0.831&0.966&0.991\\
\hline
\end{tabular}
\label{table_MAP}
\vspace{-0.3cm} 
\end{table*}

\section{Experiments}
In this section,
we evaluate the proposed method on several datasets,
and compare to the prior methods.
Moreover,
we give more discussions for the network design.

\begin{table}[tbp]
\centering
\caption{Comparisons on iBims-1 dataset.
The 1-st group is non-light weight methods.
The 2-nd group is light-weight methods.}
\label{table_iBims}
\begin{tabular}{|p{2.3cm}|p{0.6cm} p{0.6cm} p{0.6cm}|p{0.5cm} p{0.5cm} p{0.5cm}|}
\hline

Methods & REL$\downarrow$ & RMS$\downarrow$ & log10$\downarrow$ &$\delta_1\uparrow$ & $\delta_2 \uparrow$ & $\delta_3 \uparrow$ \\

\hline
\hline
Eigen \emph{et al.} \cite{Eigen} & 0.32 & 1.55 & 0.17& 0.36 & 0.65 & 0.84 \\
Eigen \emph{et al.} \cite{Eigen_2015_ICCV}& 0.25 & 1.26 & 0.13 & 0.47 & 0.78 & 0.93 \\
Laina \emph{et al.} \cite{Laina}& 0.23 & 1.20 & 0.12& 0.50 & 0.78 & 0.91  \\
Hu \emph{et al.} \cite{Revisiting}& 0.24 & 1.20 & 0.12& 0.48 & 0.81 & 0.92  \\
Chen \emph{et al.} \cite{SARPN}& 0.25 & 1.07 & 0.10 & 0.56 & 0.86 & 0.94 \\
Fu \emph{et al.} \cite{DORN_2018_CVPR}& 0.23 & 1.13 & 0.12& 0.55 & 0.81 & 0.92  \\
Lee \emph{et al.} \cite{Lee_2019_CVPR}& 0.23 & 1.09 & 0.11& 0.53 & 0.83 & 0.95  \\
Yin \emph{et al.} \cite{Yin_2019_ICCV}& 0.24 & 1.06 & 0.11 & 0.54 & 0.84 & 0.94 \\
Bhat \emph{et al.} \cite{AdaBins}& 0.21 & 0.91 & 0.10& 0.55 & 0.86 & 0.95  \\
\hline
Wofk \emph{et al.} \cite{Fastdepth}& 0.38 & 1.76 & 0.21 & 0.30 & 0.56 & 0.74 \\
Nekrasov \emph{et al.} \cite{Jointdepth}& 0.52 & 1.57 & 0.16& 0.33 & 0.66 & 0.87  \\
\hline
\hline
Ours &  0.26 & 1.11 & 0.11& 0.55 & 0.86 & 0.94  \\
\hline
\end{tabular}
\label{table_iBims}
\vspace{-0.2cm} 
\end{table}

\subsection{Dataset and Implementation Details}
\noindent
\textbf{Datasets:}
NYUDv2 \cite{NYUD2} and iBims-1 dataset \cite{iBims-1} are used to conduct experiments.
NYUDv2 \cite{NYUD2} is an indoor dataset that collects 464 scenes with 120K pairs of RGB and depth maps.
Following \cite{Revisiting,SARPN},
we train DANet on 50k images sampled from raw training data and adopt the same data augmentation strategy as \cite{SARPN}.
The test set includes 654 images with filled-in depth values. 
iBims-1 dataset \cite{iBims-1} contains 100 pairs of high-quality depth map and high-resolution image.
Since the dataset lacks training set,
we evaluate the generalization on it by using the model trained on NYUDv2 dataset.

\noindent
\textbf{Implementation Details:}
DANet is constructed on \ssf{the} Pytorch framework using a single NVIDIA 3090 GPU.
\edit{Our backbone, namely, EfficentNet b0,
is pre-trained on ILSVRC \cite{imagenet}.
Other parameters are randomly initialized.}
The Adam optimizer is adopted with parameters $(\beta_1, \beta_2) = \left(0.9,0.999\right)$.
The weight decay is $10^{-4}$. 
We train our model for 20 epochs with batch size of 24,
10 epochs for the local stage and 10 epochs for the global stage.
The initial learning rate is set to 0.0002 and reduced by 10 $\%$ for every 5 epochs.

\noindent
\textbf{Metrics:}
Following \cite{AdaBins}, we evaluate our method based on following metrics:
mean absolute relative error (REL),
root mean squared error (RMS),
mean log$_{10}$ Error (log10),
and the accurate under threshold ($\delta_k \textless 1.25^k, k=1,2,3$). 
Referring to \cite{AdaBins}, in order to make a fair comparison, we re-evaluated some methods \cite{Eigen,Eigen_2015_ICCV,Laina,DORN_2018_CVPR,Revisiting,SARPN,Yin_2019_ICCV}, in which the performance will be slightly different.










\subsection{Comparison with the prior methods}

\noindent
\textbf{Quantitative Evaluation:}
Table \ref{table1} shows the comparison between our method and the prior methods on NYUDv2 dataset.
The backbones of three non-real time networks \cite{Revisiting,Yin_2019_ICCV,SARPN} are replaced for comparison (Group \ding{174}).
DANet achieves a comparable RMS and accuracy of several non-real time networks \cite{Lee_2019_CVPR,DORN_2018_CVPR,Revisiting},
but only expending $1.4\%\sim1.56\%$ FLOPs of them.
It also outperforms all light-weight networks \cite{Fastdepth,Jointdepth,boosting} by a large margin.
Furthermore,
compared to the state-of-the-art methods with EfficientNet b0,
DANet 
gains the best performance on all metrics,
which expresses the effectiveness of distribution alignment in light-weight network. 
Although DANet uses more parameters than light-weight models,
it is much slighter than heavy-weight models,
enough to run well on the embedded platforms.

Table \ref{table_iBims} shows the cross-dataset evaluation on iBims-1 dataset by using the model trained on NYUDv2 dataset without fine-tuning.
\edit{Note that we do not re-normalize the depth range of the results to iBims-1.}
Although iBims-1 dataset has a totally different data distribution from NYUDv2 dataset,
DANet achieves the fifth best RMS and tied for $2$-rd best accuracy of $\delta_1$ with state-of-the-art methods \cite{AdaBins,DORN_2018_CVPR}.
Furthermore,
DANet exceeds the prior real-time works \cite{Jointdepth,Fastdepth} by a large margin on all metrics.
\edit{The reason is that DANet gains an outstanding performance on scenes with a similar depth range to NYUD v2,}
which proves the generalization of our method with distribution alignment.


\begin{figure*}[t]
\vspace{0.1cm} 
\centering
\includegraphics[width=1\linewidth]{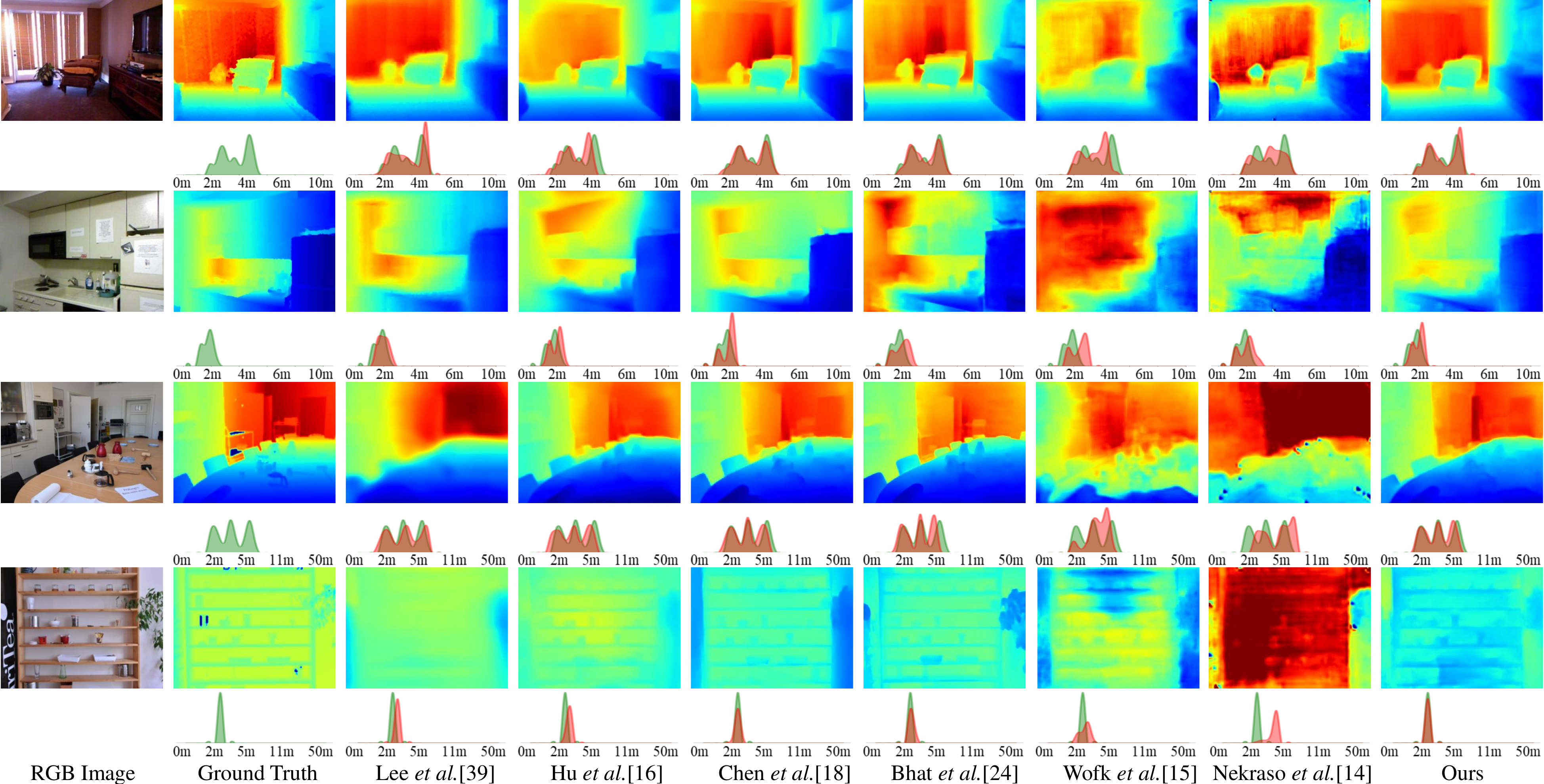}
\caption{Visualizations on the NYUDv2 (first two rows) and iBims-1 datasets (last two rows). 
  $1^{st}-2^{nd}$ columns are input and ground truth,
and $3^{th}-9^{th}$ columns for \cite{Lee_2019_CVPR,Revisiting,SARPN,AdaBins,Fastdepth,Jointdepth} and DANet.
The depth distribution is under the depth maps with green for correct depth and red for prediction.
  }
 \label{fig:NYUDv2_qualitative}
 \vspace{-0.2cm} 
\end{figure*}

\begin{table}[t]
\centering
\caption{Quantitative results of our proposed module.}
\label{table3}
\begin{tabular}{|l|c|c|c|c|}
\hline
Models & RMS$\downarrow$&$\delta_1\uparrow$&FLOPs&Params\\
\hline
\hline
Baseline & 0.510& 0.810&1.0G&3.7M \\
+ PST& 0.498 & 0.820&1.5G&8.2M\\
+ PST + \edit{min-max loss} & 0.498& 0.825&1.5G&8.2M\\
+ PST + \edit{depth-related weight} & 0.496& 0.822&1.5G&8.2M\\
+ LGO&0.496 & 0.822&1.0G&3.7M\\
+ PST + LGO & 0.488& 0.831&1.5G&8.2M  \\
\hline
\end{tabular}
\label{table_MAP}
\end{table}

\begin{table}[t]
\centering
\caption{Quantitative results of context learning module.}
\label{table4}
\begin{tabular}{|l|c|c|}
\hline
Models&RMS$\downarrow$&$\delta_1 \uparrow$\\
\hline
\hline
Ours with PPM \cite{PSPNet} & 0.497& 0.823 \\
Ours with ASPP \cite{ASPP}& 0.494& 0.825 \\
Ours with mini ViT \cite{AdaBins}& 0.496&0.822 \\
Ours with PST& 0.488& 0.831 \\
\hline
\end{tabular}
\label{table_MAP}
\vspace{-0.3cm} 
\end{table}

\begin{figure}[t]
 \centering
 \includegraphics[width=1\linewidth]{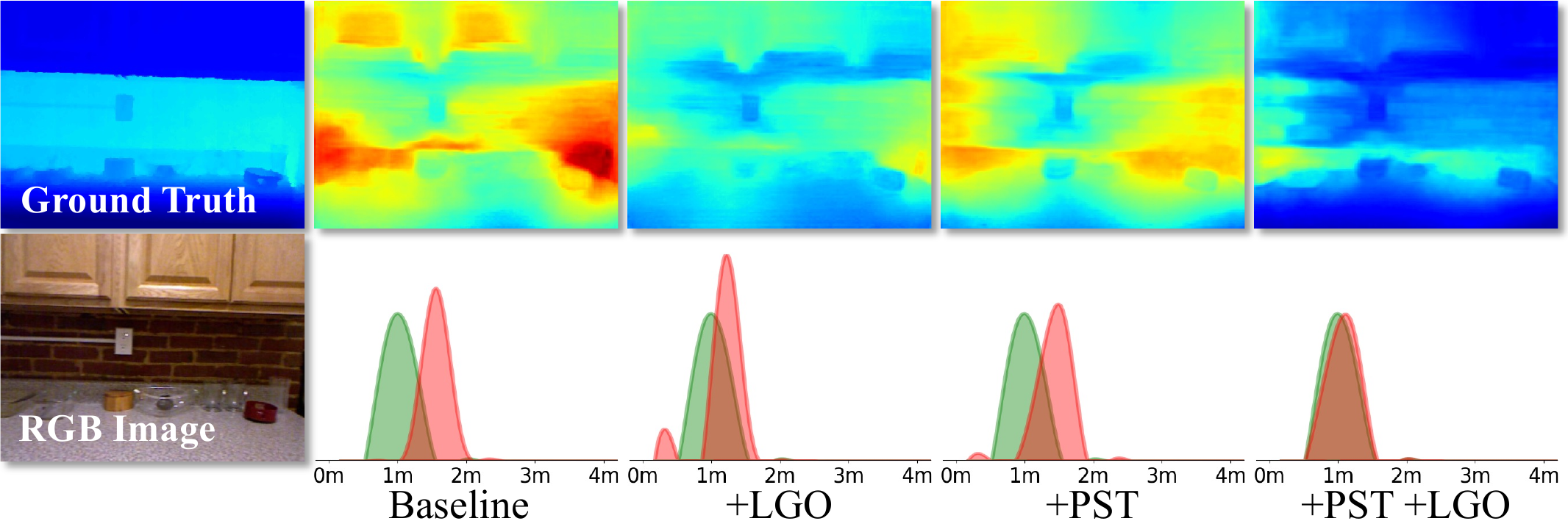}
  \caption{Qualitative results of each contributions.
  }
 \label{fig:PST_ablation}
 \vspace{-0.4cm} 
\end{figure}




\noindent
\textbf{Qualitative Evaluation:}
Fig. \ref{fig:NYUDv2_qualitative} shows the qualitative results on NYUDv2 dataset (first two rows) and iBims-1 dataset (last two rows).
In the first scene,
several methods predict a wrong depth of the wall behind the sofa,
thus suffering from the wrong depth range.
DANet gets a depth distribution almost coinciding with ground truth.
In the second scene,
the farthest region is occluded by the cabinet.
The light-weight models obtain the wrong farthest region,
causing a large distribution drift.
Our method correctly estimates the depth together with other state-of-the-art methods.
The third and fourth rows show two scenes that have never been seen.
Many methods suffer from the deviation of depth range,
especially the light-weight models \cite{Jointdepth,Fastdepth}.
Our method still estimates the depth distribution almost perfectly,
and predicts a reasonable depth image.
These visualizations further prove the effectiveness of proposed paradigm.
\subsection{Detailed Discussions}

\noindent
\textbf{Ablation studies:}
We verify our PST and LGO in Table \ref{table3}.
They are added one by one to test the effectiveness of each proposal.
Note that the baseline is an encoder-decoder network without PST and LGO.
Compared with baseline, PST and LGO achieve $1\%$ and $1.2\%$ gain in $\delta_1$.
Moreover, Baseline+PST+LGO achieves the best performance in all metrics of evaluation.
\edit{We further validate the effectiveness of min-max loss $L_{minmax}$ and depth-related weight $\lambda^{D}_i$ in LGO.
Compared to Baseline+PST,
the performance of the model is improved after using $L_{minmax}$ and $\lambda^{D}_i$ respectively.}
\vspace{-0.4cm}

Fig. \ref{fig:PST_ablation} illustrates the visualized results of these variants.
The model using LGO squeezes the depth range into a narrower space,
but fails to optimize the distribution shape.
The model using PST obtains a similar distribution with ground truth,
but suffers from the wrong depth range.
The model using all of them aligns the depth distribution well.

\begin{figure}[t]
\centering
\includegraphics[width=0.85\linewidth]{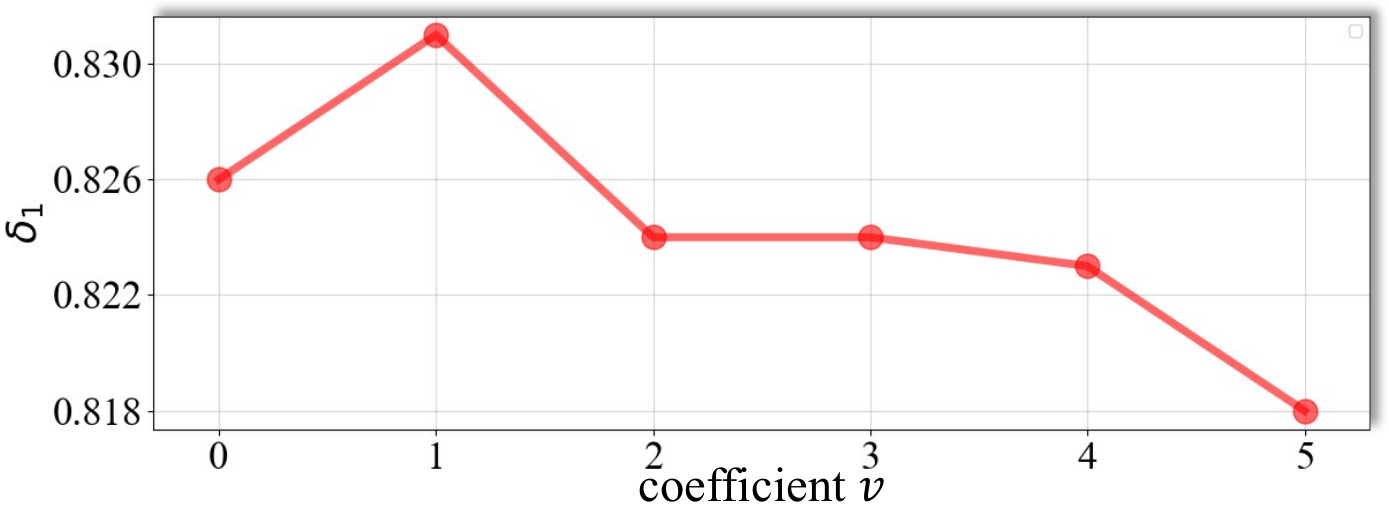}
\caption{The performance of $\delta_1$ with various coefficient $v$. } 
\label{fig:valley_b}
\vspace{-0.4cm} 
\end{figure}

\noindent
\textbf{Effectiveness of multi-scale interaction}.
To evaluate the multi-scale interaction,
PST is replaced by other contextual learning modules,
i.e., Pyramid Pooling Module (PPM) \cite{PSPNet},
ASPP \cite{ASPP}, and mini ViT \cite{AdaBins}, respectively.
As shown in Table \ref{table4},
DANet with PST outperforms others \ssf{over all} metrics, because the interaction of multi-scale regions directly models the relationship between every two regions.

\noindent
\textbf{Coefficient of depth-related weight}.
To explore the best coefficient $v$ of depth-related weight $\lambda^D_i$,
$v$ is set to $\{0, 1, 2, 3, 4, 5\}$.
Fig. \ref{fig:valley_b} demonstrates the comparisons.
It can be seen that $\delta_1$ rises at the beginning and continues to decrease as $v$ increases,
which reveals that excessive attention to the far and near areas leads to performance saturates.
Therefore, the coefficient is set to $1$ in this paper.





\section{CONCLUSIONS}
In this work, our DANet is designed to solve the distribution drift problem in light-weight MDE network.
To obtain an aligned depth distribution shape,
the PST is introduced,
which captures the interaction between multi-scale regions.
In addition,
a local-global optimization is proposed to guide the network to obtain a reliable depth range.
Experimental results on NYUDv2 and iBims-1 datasets prove that DANet achieves comparable performance with state-of-the-art methods with only 1\% FLOPs of them.
\reff{In the future,
we will further achieve real-time running time on the embedding platform,
so that it can be used to improve depth-dependent tasks \cite{2014ONLINE,ZheLiu-AAAI2019-TANet} and mobile robot applications \cite{Tiny_Obstacle2}.}


\bibliographystyle{IEEEtran}
\bibliography{IEEEfull}
\end{document}